%% file: main.tex
\documentclass{article}

\usepackage{iclr2023_conference,times}

\iclrfinalcopy %

\usepackage[utf8]{inputenc} %
\usepackage[T1]{fontenc}    %
\usepackage{hyperref}       %
\usepackage{url}            %
\usepackage{booktabs}       %
\usepackage{amsfonts}       %
\usepackage{nicefrac}       %
\usepackage{microtype}      %
\usepackage[algo2e]{algorithm2e} 
\usepackage{amsmath,amsthm}
\usepackage{mathtools}
\usepackage{amsfonts,amssymb,dsfont}
\usepackage{xfrac}
\usepackage{bm}
\usepackage{algorithm,algorithmic}
\usepackage{cleveref}
\usepackage{multirow}
\newcommand{\calD}{\mathcal D}

\newcommand{\calL}{\mathcal L}
\newcommand{\prm}{\theta}

\newcommand{\DD}{\mathcal D}

\renewcommand{\eqref}[1]{Eq. (\ref{#1})}

\DeclareMathOperator*{\softplus}{softplus}
\DeclareMathOperator*{\softmax}{softmax}

\title{Sequential Learning of Neural Networks\\for Prequential MDL}
\date{September 2021}

\author{Jorg Bornschein \\ 
\texttt{\small bornschein@deepmind.com}
\And
Yazhe Li \\
\texttt{\small yazhe@deepmind.com}
\And
Marcus Hutter \\
\texttt{\small mhutter@deepmind.com}
}

\begin{document}

\maketitle

\begin{abstract}
Minimum Description Length (MDL) provides a framework and an objective for principled model evaluation.
It formalizes Occam's Razor and can be applied to data from non-stationary sources. 
In the prequential formulation of MDL, the objective is to minimize the cumulative next-step log-loss when sequentially going through the data and using previous observations for parameter estimation.
It thus closely resembles a continual- or online-learning problem. 
In this study, we evaluate approaches for computing prequential description lengths for image classification datasets with neural networks. 
Considering the computational cost, we find that online-learning with rehearsal has favorable performance compared to the previously widely used block-wise estimation.
We propose {\em forward-calibration} to better align the models predictions with the empirical observations and introduce 
{\em replay-streams}, a minibatch incremental training technique to efficiently implement approximate random replay while 
avoiding large in-memory replay buffers. 
As a result, we present description lengths for a suite of image classification datasets that improve upon previously reported results by large margins.
\end{abstract}

\input{01-intro}

\input{02-mdl-example}

\input{03-methods}

\input{04-experiments}

\input{05-discussion}

\bibliography{main}
\bibliographystyle{iclr2023_conference}

\newpage
\appendix
\input{0X-supplement}

\end{document}

%% file: 01-intro.tex
\section{Introduction}

Within the field of deep learning, the paradigm of {\em empirical risk minimization} 
(ERM, \cite{Vapnik1991-gj}) together with model and hyper-parameter selection based on 
held-out data is the prevailing training and evaluation protocol. 
This approach has served the field well, 
supporting seamless scaling to large model and data set sizes.
The core assumptions of ERM are: a) the existence 
of fixed but unknown distributions $q(x)$ and $q(y|x)$ that represent the 
data-generating process for the problem under consideration; 
b) the goal is to obtain a function $\hat{y} = f(x, \prm^*)$ that minimizes some loss $\calL(y, \hat{y})$ in expectation 
over the data drawn from $q$; and c) that we are given a set of
(i.i.d.) samples from $q$ to use as training and validation data.
Its simplicity and well understood theoretical properties make ERM 
an attractive framework when developing learning machines. 

However, sometimes we wish to employ deep learning techniques in situations where not all these basic assumptions hold. 
For example, if we do not assume a fixed data-generating distribution $q$ we enter the realm of 
{\em continual learning}, {\em life long learning} or {\em online learning}. 
Multiple different terms are used for these scenarios because they operate under different constraints,
and because there is some ambiguity about what problem exactly a learning machine is supposed to solve. 
A recent survey on continual- and life-long learning for example describes multiple, sometimes conflicting desiderata
considered in the literature: forward-transfer, backward-transfer, avoiding catastrophic forgetting and maintaining 
plasticity \citep{Hadsell2020-fn}. 

Another situation in which the ERM framework is not necessarily
the best approach is when minimizing the expected loss 
$\calL$ is not the only, or maybe not even the primary objective of the learning machine. 
For example, recently deep-learning techniques have been used to aid structural and causal 
inference \citep{vowels2021d}. In these cases, we are more interested in model selection 
or some aspect of the learned parameters $\prm^*$ than 
the generalization loss $\calL$. 
Here we have little to gain from the generalization bounds provided by the ERM framework 
and in fact some of its properties can be harmful.

Independent of ERM, compression based approaches to inference and learning such as Minimum Description Length \citep{Rissanen1984,Grunwald2004}, Minimum Message Length \citep{Wallace2005} and 
Solomonoffs theory of inductive inference \citep{solomonoff1964formal,Hutter:11uiphil} have been extensively studied.
In practice, however, they have been primarily applied to small scale problems and with very simple models compared to deep neural networks.
These approaches strive to find and train models that can compress the observed data well and rely on the fact that such models 
have a good chance of generalizing on future data.
It is considered a major benefit that these approaches come with a clear objective and have been studied 
in detail even for concrete sequence of observations; without assuming stationarity or a probabilistic generative process at all
(also called the {\em individual sequence} scenario). 
The individual sequence setting is problematic for ERM because the fundamental step of creating 
training and validation splits is not clearly defined. 
In fact, creating such splits implies making an assumption about what is equally distributed 
across training and validation data. %
But creating validation splits can be problematic even with (assumed) stationary distributions, 
and this becomes more pressing as ML research moves from curated academic benchmark data sets to large user provided or web-scraped data.
We discuss examples in \Cref{sec:splitting}.
In contrast to ERM, compression approaches furthermore include a form of Occam's Razor, 
i.e. they prefer simpler models (according to some implicit measure of complexity) as long 
as the simplicity does not harm the model's predictive performance.  

The literature considers multiple, subtly different approaches to defining and computing description lengths. 
Most of them are intractable and known approximations 
break down for overparameterized model families such as neural networks.
One particular approach however, prequential MDL \citep{dawid1999prequential,hutter2005asymptotics},
turns computing the description length $L(\DD | M) = - \log p(\DD | M)$ into a specific kind of continual or online learning problem: 
$\log p(\DD | M){:=}\sum_{t=1}^T \log p_M(y_t | x_t, \hat{\prm}(\DD_{<t}))$, where $\DD{=}\{(x_t, y_t)\}_1^T$ is the sequence of inputs 
$x_t$ and associated prediction targets $y_t$; $p_M(y | x, \prm)$ denotes a model family $M$ and $\hat{\prm}(\DD_{<t})$ an 
associated parameter estimator given training data $\DD_{<t}$.
In contrast to ERM which considers the performance of a learning algorithm on held-out data after training on a fixed dataset,
prequential MDL evaluates a learner by its generalization performance on $y_t$ after trained on initially short
but progressively longer sequences of observations $\DD_{<t}$. This resembles an online learning problem where a learner is sequentially exposed to new data $(x_t, y_t)$, 
however a sequential learner is also allowed to revisit old data $x_{t'}, y_{t'}$ for training and when making a prediction at time $t{>}t'$.

\paragraph{Contributions.}
\label{sec:contributions}

Previous work on computing prequential description lengths with neural networks relied on 
a block-wise (chunk-incremental) approximation: at some positions $t$ a model is trained
from random initialization to convergence on data $\DD_{<t}$ and then their prediction losses 
on the next intervals are combined \citep{Blier2018-rl,bornschein2020small,jin2021causal,bornschein2021causal,perez2021rissanen,whitney2020evaluating}.
We investigate alternatives that are inspired by continuous-learning (CL) based methods. In particular,
chunk-incremental and mini-batch incremental fine-tuning with rehearsal. 
Throughout this empirical study, we consider the computational costs associated with these methods. 

We propose two new techniques: {\em forward-calibration} and {\em replay streams}. 
{\em Forward-calibration} improves the results by making the model's predictive uncertainty better match the observed distribution. 
{\em Replay streams} makes replay for mini-batch incremental learning 
more scalable by providing approximate random rehearsal while avoiding large in-memory replay-buffers.
We identify exponential moving parameter averages, label-smoothing and weight-standardization as generally 
useful techniques.
As a result we present description lengths for a suite of 
popular image classification datasets that improve upon the 
previously reported results \citep{Blier2018-rl,bornschein2020small} by large margins.

The motivation for studying prequential MDL stems from the desire to apply deep-learning techniques 
in situations violating the assumptions of ERM. In this work we concentrate on established 
i.i.d. datasets. While the motivation for choosing MDL over ERM is less clear in this scenario, it allows us 
to evaluate well known architectures and compare some results from ERM-based training, 
which has been the focus of the community for decades. We include additional experimental 
results on non-stationary data in Appendix \ref{sec:cloc}.

%% file: 02-mdl-example.tex
\begin{figure}[t]
\includegraphics[width=0.18\linewidth]{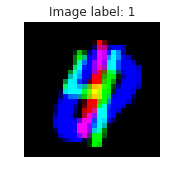}
\includegraphics[width=0.42\linewidth]{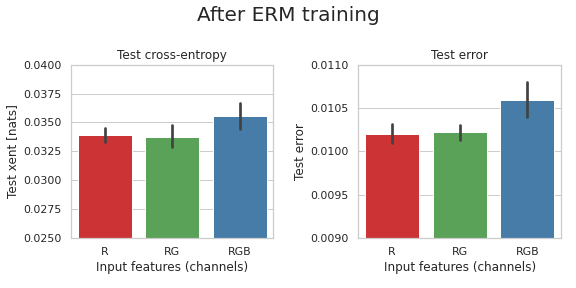}
\includegraphics[width=0.42\linewidth]{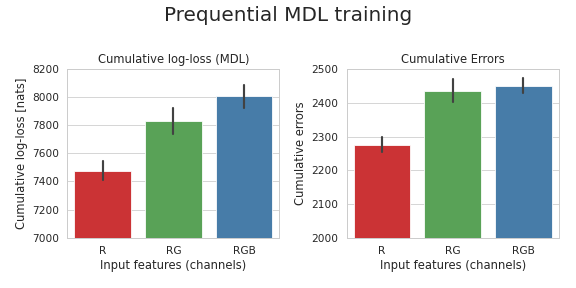}
\vspace{-0.6cm}
\caption{ 
{\bf Left:} Example images. 
{\bf Center:} Best performing ERM-trained models on held-out data.
There is no clear indication which conditioning should be preferred.
{\bf Right:}  Log-loss and errors accumulated throughout the sequence. We observe
that MDL based model selection provides strong evidence that the model conditioned on the R-channel alone should be preferred. 
In all cases we randomly sampled 50 hyperparameters and use bootstrap-sampling to obtain 95\% confidence intervals for the 
performance of the best performing model. See \Cref{fig:rgbmnist-regret} for detailed regret plots.
}
\label{fig:rgbmnist-results}
\end{figure}
\begin{figure}[t]
\includegraphics[width=0.99\linewidth]{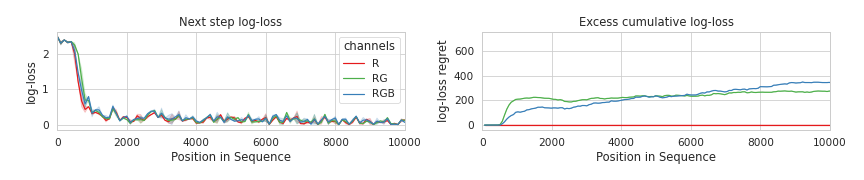}
\vspace{-0.6cm}
\caption{
{\bf Left:} Next-step neg. log-loss for the best performing models
as a function of the position. The models are trained on all the previous examples. We observe that the models 
conditioned on the R, R and G, or R, G and B channels all rapidly improve their predictive performance when 
trained on more data. 
{\bf Right:} The cumulative neg. log-loss (= description length) 
relative to the best performing model which is conditioned 
on the R channel only. During the first $\approx$ 2k examples 
the models conditioned on irrelevant channels accumulate hundreds of excess nats description length.
\label{fig:rgbmnist-regret}
}
\end{figure}

\paragraph{An example: model selection using ERM v.s.\ MDL.}
\label{sec:rgbmnist}
To showcase the properties of prequential MDL and its bias towards simpler models, we 
generate images by randomly drawing 3 MNIST examples and superimposing them
as red(R), green(G) and blue(B) channels. The label is always associated with the red channel; 
green and blue are distractors.
We train a LeNet \citep{lecun1989} convolutional neural network on 100k examples from this dataset; either
within the ERM framework by drawing i.i.d. examples, training to convergence and validating on held-out data, 
or within the MDL framework by treating the data as a sequence, training online with replay buffer (see \Cref{sec:replay} for details) and recording the (cumulative) next-step prediction losses and errors.
We compare 3 different models: a model conditioned only on the R channel, on the R and G, or on R, G and B channels.
\Cref{fig:rgbmnist-results} (middle) shows the validation-set performance of the best performing models after ERM training. \Cref{fig:rgbmnist-results} (right) shows description lengths on the same data.
It is evident that ERM does not give a clear indication which 
conditioning should be preferred. However, MDL strongly prefers the model conditioned  
only on the R-channel.
\Cref{fig:rgbmnist-regret} shows a more detailed analysis of the three differently conditioned models in the 
MDL scenario. We observe that prequential MDL identifies the minimal conditioning because it 
takes the small-data performance of the models into account. Given sufficient training
data the three alternatives perform almost indistinguishable (as seen from their almost parallel regret curves).
When we treat the description lengths as evidence for Bayesian model selection to choose between 
$M_\text{R}, M_\text{RG}$ and $M_\text{RGB}$ we obtain
$p(M_\text{R} | \DD) = \sfrac{p(\DD | M_\text{R})}{p(\DD | M_\text{R}) + p(\DD | M_\text{RG}) + p(\DD | M_\text{RGB})}$.
With the description length $-\log p(\DD | M_\text{R})$ about 300 nats smaller than the ones for $M_\text{RG}$ and $M_\text{RGB}$, 
we have $\approx 1 {:} e^{-300}$ odds in favour of $M_\text{R}$.
The \Cref{sec:rgbmnist-appendix} contains additional experiments with more architectures and 
the experimental details; demonstrating that this is a robust result regardless of architectures 
and optimization hyperparameters.


%% file: 03-methods.tex
\section{Online Learning Approaches to Prequential MDL Estimation}
\label{sec:methods}

\begin{algorithm}[t]
\caption{Mini-batch Incremental Training with Replay Streams}
\begin{algorithmic}[1]
  \REQUIRE data $(x_t, y_t)_{t=1}^T$; augmentation $Aug(.)$; number of replay streams $K$; EMA decay $\alpha$ 
  \STATE Initialize: parameters $\theta$; EMA parameters $\bar{\theta} = \theta$; softmax temperature calibration $\beta = 1$; \\ Replay positions $\{\rho_k = 1 \}_{k=1}^K$; $\mathcal{L}_{preq}=0$
  \FOR{$t = 1$ \TO $T$}
    \STATE Compute next-step loss: $\mathcal{L}_{t} := -\log p(y_t | x_t, \bar{\theta}, \beta)$
    \STATE Update cumulative loss: $\mathcal{L}_{preq} \leftarrow \mathcal{L}_{preq} + \mathcal{L}_{t}$
    \STATE Update temperature calibration: $\beta \leftarrow \beta - \nabla_\beta \log p(y_t | x_t, \bar{\theta}, \beta)$
    \STATE Update parameters: $\theta \leftarrow \theta - \nabla_\theta \log p(y_t | \tilde{x}_t, \theta, 1.)$, with $\tilde{x}_t = Aug(x_t)$
  \STATE Update EMA parameters: $\bar{\theta}  \leftarrow (1 - \alpha) \bar{\theta} + \alpha \, \theta$
    \FOR{$k = 1$ \TO $K$} 
      \STATE Get data from stream $k$: $x \leftarrow x_{\rho_k}$; $y \leftarrow y_{\rho_k}$, advance position $\rho_k \leftarrow \rho_k + 1$
      \STATE Update parameters: $\theta \leftarrow \theta - \nabla_\theta \log p(y | \tilde{x}, \theta, 1.)$ with $\tilde{x} = Aug(x)$
      \STATE Update EMA parameters: $\bar{\theta} \leftarrow (1 - \alpha) \bar{\theta} + \alpha \, \theta$
      \STATE Maintain replay distribution: with probability $p_\text{reset}$ reset $\rho_k \leftarrow 1$ (see  \Cref{eq:reset-prob})
    \ENDFOR
  \ENDFOR
  \RETURN $\mathcal{L}_{preq}$
\end{algorithmic}
\label{alg:replay-stream}
In practice we perform the algorithm with mini-batches instead of individual examples 
and use gradient based optimizers such as AdamW\citep{loshchilov2018decoupled} instead of plain SGD.
\end{algorithm}

In this section, we describe practical strategies for computing 
prequential description lengths with neural networks.
Conceptually, computing the cumulative log-loss resembles a 
continual- or online-learning problem without constraints such as limiting access to previously observed data.
In practice, however, some naive approaches such as retraining from 
scratch to convergence whenever a new example arrives are infeasible due to their 
high resource usage. 
In our study, we consider the compute requirements of each approach by counting the 
number of floating-point operations (FLOPs) until the learner reaches the end of the data sequence.
We do not limit the learner's ability to access or sample previously observed data because storing and reading from large datasets is generally not considered a major technical bottleneck.
However, in-memory (RAM) based implementations of replay-buffers for continual learning (rehearsal) suffer from limited capacity because data sets for training deep neural networks often exceed RAM sizes. 
Below we describe {\em replay-streams}, a simple approach to approximate random sampling for replay while only utilizing 
cheap sequential access to data on disk. With this approach, unlimited 
approximate random rehearsal from data on permanent storage becomes as easy as sampling 
from large datasets for ERM based optimization.

We compare the following approaches to compute prequential description lengths:

\paragraph{Chunk Incremental / From-Scratch (CI/FS):} 
Variants of this approach have been used to compute the recently reported description lengths 
for deep-learning models 
\citep{Blier2018-rl,bornschein2020small,jin2021causal,bornschein2021causal, perez2021rissanen}.
The data-sequence is partitioned into $K$ non-overlapping intervals; typically of 
increasing size (i.e. choosing exponentially spaced split points 
$\{s_k\}_{k=1}^K$, with $s_k \in [2, \dots, n]$, $s_k{<}s_{k+1}$ and $s_K{=}N$). 
For each $k$, a neural network is randomly initialized and trained to convergence on all the 
examples before the split point $s_k$ and evaluated on the data between 
$s_k$ and $s_{k+1}$. This corresponds to the block-wise approximation of the area under the curve:
$\sum_{i=1}^N \log p(y_i | \hat{\prm}(\calD_{<i})) \approx
\sum_{k=1}^{K-1} \sum_{j=s_k}^{s_{k+1}} \log p(y_j| \hat{\prm}(\calD_{<s_k}))$,
where $\hat{\prm}(\calD)$ denotes parameters after training on data $\calD$.
To ensure that the model produces calibrated predictions even when $\calD_{<s_k}$ 
is small, we use softmax temperature calibration \citep{guo2017}:
at each stage we split the data $\calD_{<s_k}$ into a 90\% training and a 10\% calibration data. 
Conceptually, we could perform post-calibration by first training the network to convergence and then, with all parameters frozen, replacing the output 
layer $\softmax(h)$ with the calibrated output layer $\softmax(\softplus(\beta) h)$, where $\beta$ is a scalar parameter chosen to minimize the loss on calibration data. 
In practice, we optimize $\beta$ by gradient descent in parallel with the other model parameters. We alternate computing gradient steps for $\theta$, calculated from the training set and using the uncalibrated network (with final layer $\softmax(h)$), with a gradient step on $\beta$, calculated from the validation set using the calibrated network (with final layer $\softmax(\softplus(\beta) h)$). This simple calibration procedure has proven to be surprisingly effective at avoiding overfitting symptoms when training large neural networks on small datasets \citep{guo2017,bornschein2020small}.

\paragraph{Chunk Incremental / Continual Fine-tuning (CI/CF):} 
Similar to CI/FS, the sequence is split into increasingly larger chunks. But instead of training
the model from scratch at each stage, the network is continuously fine-tuned on the now larger dataset $\DD_{<s_k}$.
We use the same calibration strategy as for CI/FS. We expect to save
compute by avoiding training from scratch. However, recent 
research suggests that, when first trained on data $\DD_A$ and then trained 
on data $\DD_A \cup \DD_B$, deep-learning models struggle to converge to a solution that generalizes as well as training on $\DD_A \cup \DD_B$ from 
scratch; even when both $\DD_A$ and $\DD_B$ are sampled from the same 
underlying distribution.
In \cite{ash2020warm} the authors observe that shrinking the model-parameters 
and adding some noise after training on $\DD_A$ can improve the final generalization 
performance. %
We therefore run an ablation study and perform {\em shrink \& perturb} 
operation whenever advancing to the next chunk $k$.

\paragraph{Mini-batch Incremental / Replay-Buffer (MI/RB)}
\label{sec:replay}
This approach uses continual online mini-batch gradient descent with an additional replay buffer to store previously seen examples. At each time $t$, the learner performs a number of learning steps on data in the replay buffer. This could be all or a subset of the data $\DD_{<t}$ depending on the capacity of the replay. 
We propose {\em forward-calibration} to optimize the calibration parameter $\beta$: each new batch of examples is first used for evaluation, 
then used to perform a single gradient step to optimize $\beta$, 
and finally used for training the parameters $\prm$ and placed into the replay-buffer. 
Calibration is computationally almost free: the forward-pass for 
the calibration gradient step can be shared with the evaluation forward-pass, 
and backward-propagation for $\beta$ ends right after the softmax layer.
Appendix \ref{sec:forward-calibration} contains a more detailed description.
If the replay buffer is limited by the memory capacity, we either use a FIFO replacement-strategy or reservoir-sampling \citep{vitter1985} to maintain a fixed maximum buffer size.

\paragraph{Mini-batch Incremental / Replay-Streams (MI/RS)}
\label{sec:replay-stream}

Implementing large replay buffers can be technically challenging: in-memory (RAM) replay buffers have limited capacity
and random access and sampling from permanent storage is often slow.
We propose a simple yet effective alternative to storing replay data in-memory. 
We instead assume that the data is stored in its original order on permanent storage and
we maintain $K$ replay streams (pointers into the sequence) , 
each reading the data in-order from the sequence  $(x_1, y_1) \dots (x_T, y_T)$. We can think of a replay-stream as a file-handle, currently at 
position $\rho \in 1{\dots}T$. A read operation yields $(x_\rho, y_\rho)$ and increments the position to 
$\rho \leftarrow \rho + 1$. We denote the position of the $k$th replay-stream with $\rho_k$. 
Every time the learner steps forward from time $t$ to $\tilde{t}{=}t{+}1$ we also 
read a replay example from each of the $K$ replay-streams and perform gradient steps on them. 
By stochastically resetting individual streams with probability $p_{\text{reset}}$
to position $\rho \leftarrow 1$, we can influence the distribution of positions $\rho$.
If we wish to use uniform replay of previous examples we reset with probability 
$p_\text{reset} = \sfrac{1}{\tilde{t}}$. If instead we wish to prioritize recent examples
and replay them according to $p_\text{replay}(a)$, where $a{=}\tilde{t}{-}i$ is the age of 
the example $(x_i, y_i$) relative to the current position of the learner $\tilde{t}$, we use
\vspace{-0.2cm}
\begin{align}
  \label{eq:reset-prob}
  p_{\text{reset}}(\tilde{t}) = \frac{
        \sum_{a=t}^{\tilde{t}} p_{\text{replay}}(a) 
    }{
        \sum_{a'=0}^{\tilde{t}} p_{\text{replay}}(a')
    }
\end{align}
Note that $p_{\text{replay}}$ can be an unnormalized distribution and that by setting $p_{\text{replay}} \propto 1$ 
we recover $p_\text{reset} = \sfrac{1}{\tilde{t}}$ for uniform replay.
See \Cref{alg:replay-stream} for details.

%% file: 04-experiments.tex
\section{Experiments}
\label{sec:experiments}

We empirically evaluate the approaches from \Cref{sec:methods} on a suite of models and image classification data sets. We use MNIST \citep{lecun2010mnist}, EMNIST, CIFAR-10, CIFAR-100 \citep{Krizhevsky09learningmultiple} and 
ImageNet \citep{ILSVRC15} and randomly shuffle each into a fixed sequence of examples. 
With MDL, we do not require validation or test set and could merge it with the training data. 
But because the data is known to be i.i.d. and we compare results with established ERM-based training,
we instead maintain the standard test sets as held-out data. It must be emphasized that the primary objective
throughout this study is to minimize the description length. All hyperparameters are tuned towards that goal. 
When we report test set performance it should be considered a sanity check.
We evaluate the following architectures: MLPs, VGG \citep{Simonyan15}, 
ResNets \citep{he2016deep}, WideResNets \citep{ZagoruykoKomodakis2016} and the transformer-based DeiT \citep{pmlr-v139-touvron21a} architecture.
We refer to a variant of the VGG architecture with weight-standardized convolutions \citep{weightstandardization} 
and with batch normalization \citep{ioffe2015batch} as {\em VGG++}.
For the bulk of our experiments, we use the AdamW\citep{loshchilov2018decoupled} as optimizer; a series of control experiments however suggest that the quantitative results generalize to optimizers such as RMSProp and simple mini-batch SGD with momentum.
We identified the following techniques that consistently improve the results and use them throughout this paper if not mentioned otherwise:

{\bf Exponentially moving average (EMA)} Similar to Polyak averaging \citep{polyak1992}, EMA maintains a moving average 
of the model parameters $\prm$ and is used for next-step evaluations as well as when tuning the 
softmax temperature $\beta$. %

{\bf Label smoothing} \citep{DBLP:journals/corr/SzegedyVISW15} A reliable regularizer, label smoothing often helps for the 
continual fine-tuning methods such as MI/RB, MI/RS and CI/CF.

{\bf Weight standardization} For the VGG architecture,  
weight standardized convolutions lead to significant improvements - bigger than for conventional ERM-based training. For other architectures such as ResNets, the results are not conclusive.

\begin{figure}[t]
\includegraphics[width=\linewidth]{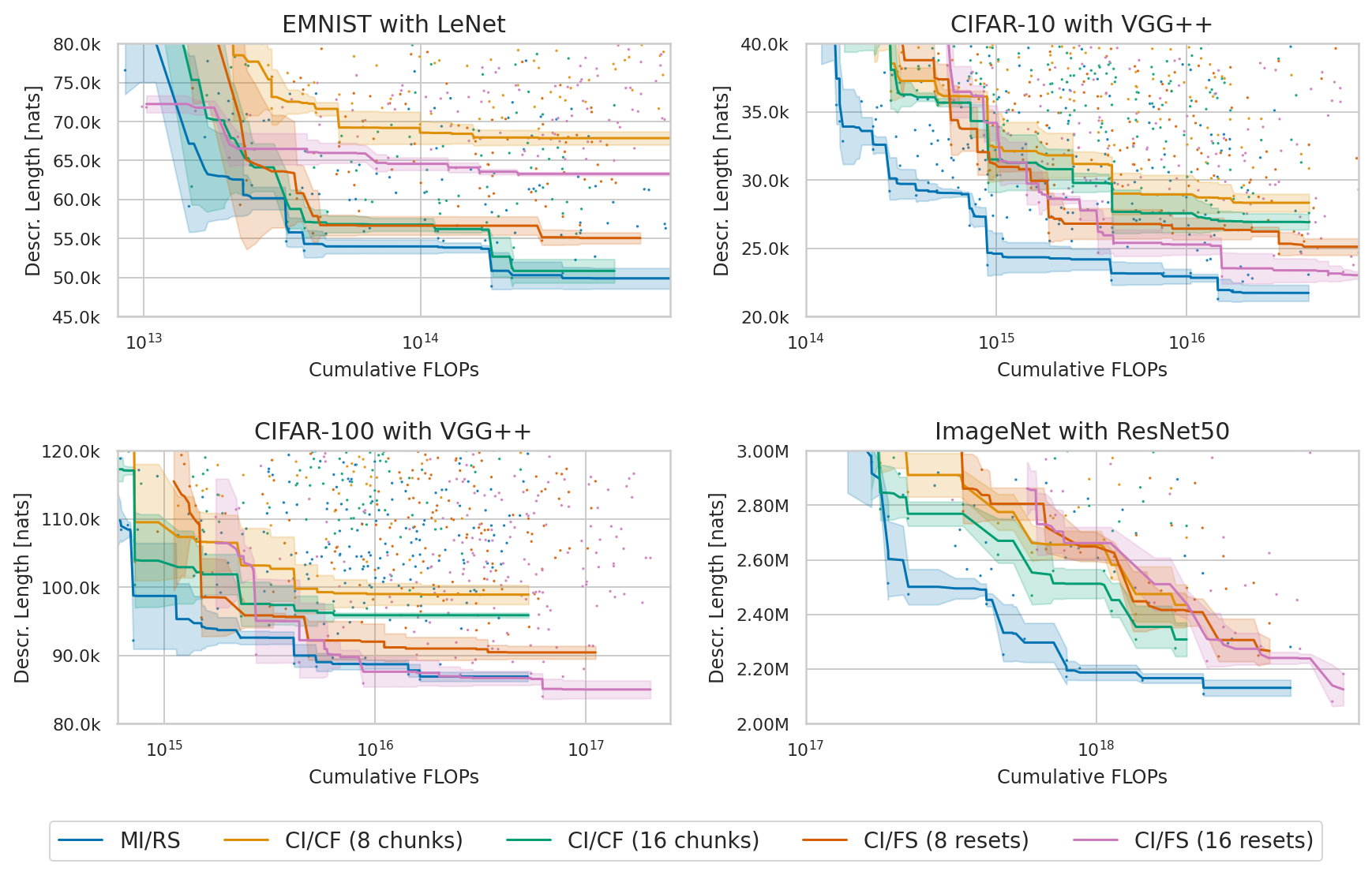}\\
\vspace{-0.5cm}
\caption{
Shortest description lengths as function of compute resources consumed for different methods.
For each method (in different colors) we run 250 (100 for ImageNet) independent experiments with 
randomly sampled hyperparameters. 
Each dot represents such an experiment and the solid lines show the Pareto-front of the best archived 
performance for different compute budgets.
With the exception of CIFAR-100 in the large-FLOPs regime, the mini-batch incremental appoach MI/RS 
results in the shortest description lengths.
}
\label{fig:pareto}
\end{figure}

\subsection{Comparison of Online Learning Approaches.}
To compare the effectiveness of the different approaches listed in \Cref{sec:methods}, we run a broad random 
hyperparameter sweep for EMNIST with LeNet, CIFAR-10 and CIFAR-100 with VGG++ and ImageNet with
ResNet50. %
The hyperparameter intervals depend on the data and are detailed in \Cref{sec:experiment-details}. 
We sample learning rate, EMA step size, batch size, weight decay; 
but crucially also number of epochs (or, correspondingly, number of replay streams for MI/RS) 
and an overall scaling of the model width (number of channels). %
We thus cover a large range of computational budgets, spanning about 2 orders of magnitude.
When the capacity of the replay-buffer matches the total length of the sequence, MI/RB and MI/RS are equivalent. 
In these cases we omit MI/RB and instead use MI/RS to represent online-learning with unlimited rehearsal. 
A study of replay buffer v.s. replay streams can be found in \Cref{sec:buffer_streams}. 
We show the overall Pareto front for these experiments in \autoref{fig:pareto}.
Mini-batch incremental learning obtains the lowest description lengths for a wide range of computational budgets. 
Only for very large compute budgets on CIFAR-100 we observe that the block-wise approximation with training from 
scratch (CI/FS) has a better results. 

\begin{table}[t]
    \centering
    \smaller
    \vspace{-.55cm}
    \begin{tabular}{llcccccccc}
         &  & \multicolumn{5}{c}{\bf MI/RS} & \multicolumn{3}{c}{\bf ERM}
         \\\cmidrule(lr){3-7}\cmidrule(lr){8-10}
         \multirow{2}{*}[-2pt]{\bf Data} & \multirow{2}{*}[-2pt]{\bf Model} & \multicolumn{2}{c}{{\bf Cumulative}} & \multicolumn{2}{c}{{\bf Test}} & & \multicolumn{2}{c}{{\bf Test}} &
         \\
         \cmidrule(lr){3-4}\cmidrule(lr){5-6}\cmidrule(lr){8-9}
        
        & & Loss & Errors & Loss & Error(\%) &  FLOPs & Loss & Error(\%) & FLOPs \\
        \hline
        MNIST & LeNet   & 4.4k & 1362 & 0.03 & 1.0 & 1.4e13 & 0.03 & 0.8 & 1.5e13 \\
        MNIST & MnistNet & 2.1k & 643 & 0.02 & 0.5  & 4.5e14 & 0.02 & 0.5 & 3.2e14  \\ 
        CIFAR-10 & VGG++ & 22.1k & 6.6k & 0.24 & 5.8  & 5.9e15 & 0.21 & 6.9 & 8.0e15 \\ 
        CIFAR-10 & WRN-28-10 & 22.9k & 7.5k & 0.23 & 7.4 & 3.7e16 & 0.20 & 6.7 & 5.4e16 \\ 
        CIFAR-100 & VGG++ & 93.7k & 23.1k & 1.14 & 32.0  & 4.2e15 & 1.14 & 28.1 & 2.6e15 \\ 
        CIFAR-100 & WRN-28-10 & 87.0k & 22.3k & 1.10 & 30.3 & 2.5e16 & 1.08 & 28.3 & 4.3e16 \\ 
        ImageNet & VGG++ & 2.55M & 549k & 1.63 & 36.8 & 7.5e18 & &  & \\ 
        ImageNet & ResNet-34 & 2.35M & 507k & 1.30 & 30.9 & 1.6e18 &1.25 & 29.9 & 7.0e17 \\   
        ImageNet & ResNet-50 & 2.21M & 496k & 1.29 & 29.1 & 8.5e17 & 1.15 & 28.7 & 1.9e18 \\ 
        ImageNet & ResNet-50$^\dagger$ & 1.93M & 431k & 1.06 & 26.1 & 4.0e18 & & & \\ 
        ImageNet & ResNet-101 & 1.97M & 447k & 1.20 & 28.2 & 3.1e18 & 1.10 & 26.6 & 4.6e18 \\ 
        ImageNet & DeiT-S & 2.73M & 588k & 1.40 & 31.7 & 1.2e18 & 1.22 & 28.1  & 4.8e18 \\ 
        \midrule
        \multicolumn{2}{l}{\textit{Prior Works:}}    & & & & & & \\
        \midrule
        MNIST\citep{Blier2018-rl} & MnistNet & 2.8k & & & 0.5 & & & & \\
        CIFAR-10\citep{Blier2018-rl} & $\approx$ VGG++ & 31.4k & & & 6.7 & & & & \\
        ImageNet\citep{bornschein2020small} & ResNet-50 & 3.32M& & & - & & & & \\
        \bottomrule
    \end{tabular}
    \vspace{-0.35cm}
    \caption{
        {\bf Left half of the table}: Suite of results when applying MI/RS for description length estimation. 
        Cummulative loss in nats; cumulative error counts the number 
        of prediction errors after reaching the end of the sequence. The test set metrics report the 
        performance of the model selected by lowest description length.
        {\bf Right half}: Results when using the same architecture and hyperparameter-sweep for 
        conventional ERM training.
        $^\dagger$ Manually tuned hyper parameters. 
    }
    \label{tab:all_results}
\end{table}

\paragraph{MDL for Deep Neural Networks.}
We apply MI/RS to a wider range of architectures. For most experiments we again use random hyperparameter 
sweeps, this time however without scaling the size of the models and with an increased lower bound for the number of 
replay streams (see \autoref{sec:experiment-details}). 
We present our results for the best run in each sweep and contrast them with previously reported description lengths 
from the literature in \Cref{tab:all_results}. 
We also show the test set performance of these models after they reached 
the end of training data.
The rightmost columns show the test set performance when performing 
the same hyperparameter sweep for conventional ERM training on the full data set.
Note that these runs do not use any kind of learning rate annealing, which is otherwise 
often employed to improve results.

\begin{figure}[h]
    \centering
    \includegraphics[width=0.48\linewidth]{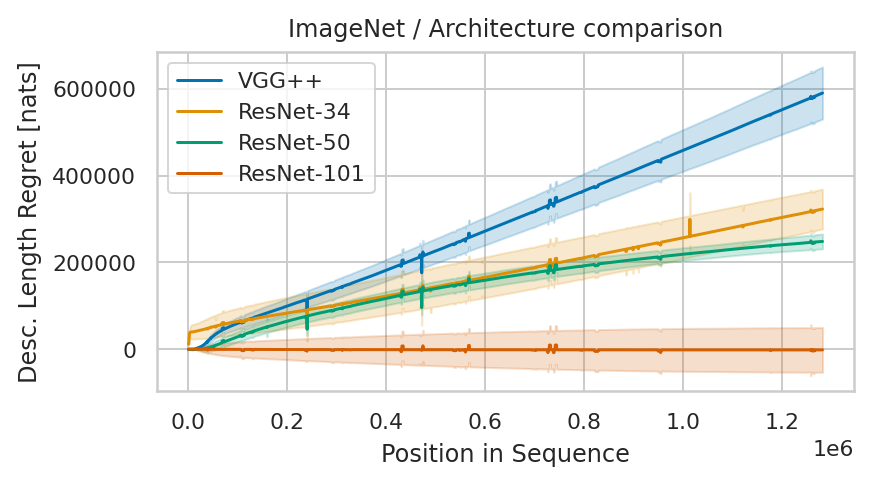}
    \includegraphics[width=0.48\linewidth]{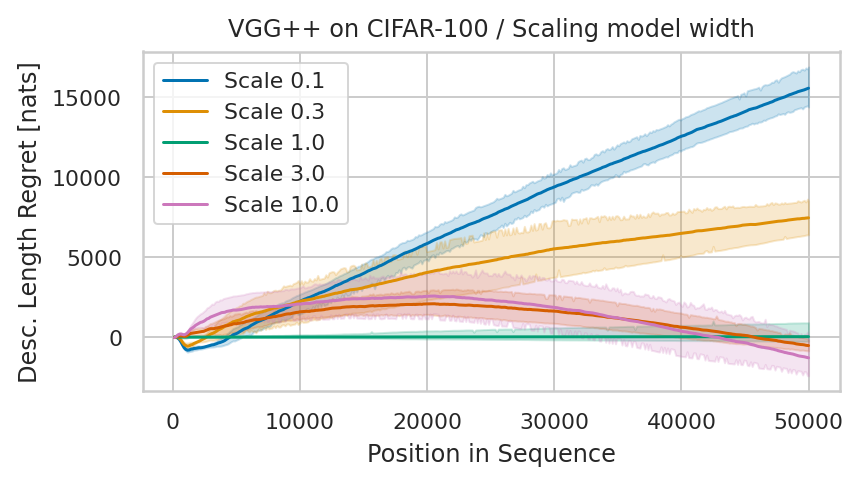} \\
    \vspace{-0.3cm}
    \caption{
    {\bf Left}: Comparing architectures on ImageNet.
    {\bf Right}: Scaling the size of VGG++. %
    }
    \label{fig:model-scaling}
\end{figure}

Based on these experiments we plot a regret comparison for different architectures trained
on ImageNet in \autoref{fig:model-scaling} (left). We observe that the additional depth 
and capacity of a ResNet101 is beneficial from the very beginning of the data-sequence.
\autoref{fig:model-scaling} (right) shows the effect of scaling the number of channels of a 
VGG++ architecture on CIFAR-100 and it is evident that decreasing the model width has negaive 
effects throughout the sequence and not just when the data gets larger.

\begin{figure}[b]
    \centering
    \includegraphics[width=0.48\linewidth]{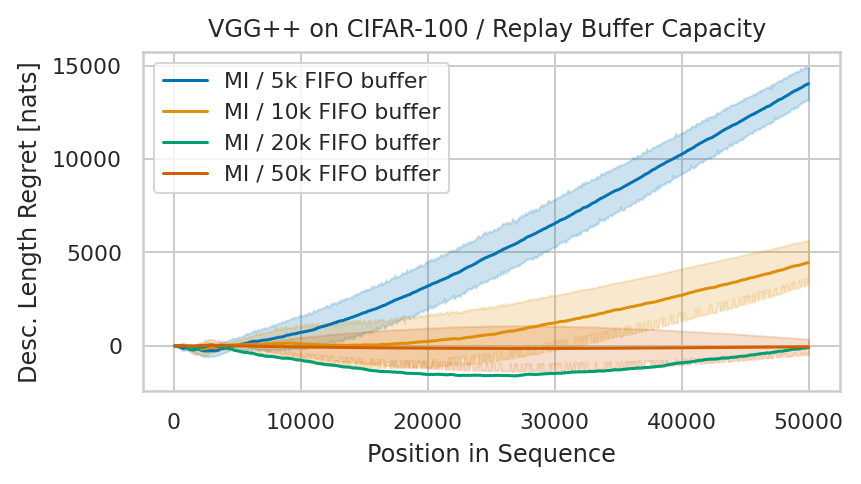}
    \includegraphics[width=0.48\linewidth]{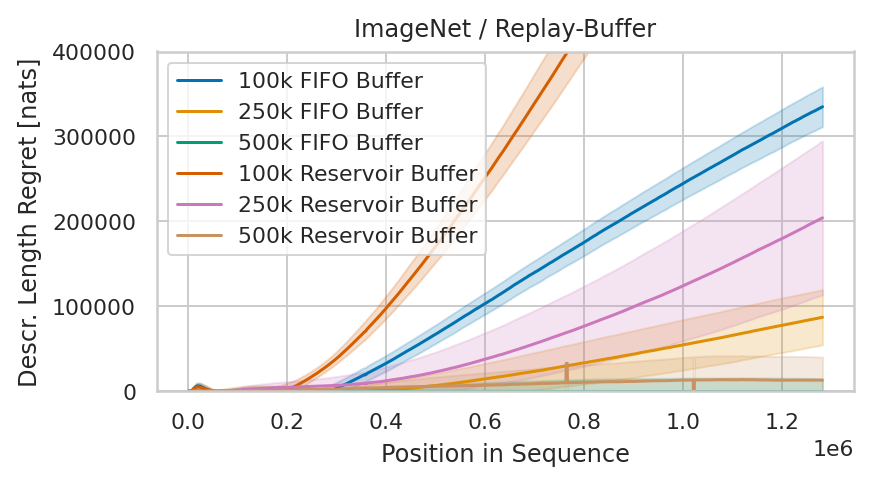} \\
    \vspace{-0.3cm}
    \caption{
    The effect of reducing the replay-buffer capacity.
    {\bf Left}: Regret for different replay buffer sizes for CIFAR-100.
    {\bf Right}: Comparing buffer capacity and FIFO vs. reservoir sampling based 
        replacement strategies for ImageNet with ResNet50.
    }
    \label{fig:imagenet-buffersize}
\end{figure}

\paragraph{Replay Buffer and Replay Streams}
\label{sec:buffer_streams}
are equivalent for short data sequences, they differ when the length of a sequence exceeds the capacity of the replay buffer. 
In this section, we study the effects of limiting the replay buffer size. 
\Cref{fig:imagenet-buffersize} shows that reducing the capacity has a significant negative effect on the results. 
In addition, the strategy of which samples to keep in the buffer plays an important role.
Using reservoir sampling instead of a FIFO replacement policy leads to further severely 
degraded results. This is maybe not surprising because with reservoir-sampling insertion to the 
buffer is stochastic; a fraction of the examples are thus never replayed at all. 
Appendix \ref{sec:replay-streams-ablations} contains ablations for the number
of replay-streams on CIFAR-100 and ImageNet.

\paragraph{Ablations.}
We provide ablations for using forward-calibration, label smoothing and weight standardization in \Cref{tab:ablations}.  %
It is evident that each of these techniques helps obtaining better description lengths for the datasets and 
architectures under consideration. Regret plots for these ablations can be found in \autoref{sec:more-experiments}.

\begin{table}[h]
\vspace{0.2cm}
\begin{center}
\small
\begin{tabular}{llcr@{\hskip2pt}lr@{\hskip2pt}lr@{\hskip2pt}lr@{\hskip2pt}l}\toprule
    {\bf Data } & {\bf Model} & {\bf None}& \multicolumn{2}{c}{\bf +FC} & \multicolumn{2}{c}{\bf +LS} 
                              & \multicolumn{2}{c}{\bf +WS} & \multicolumn{2}{c}{\bf +ALL} \\
    \hline
    CIFAR-10 & VGG++ & 32.4k & 31.8k & (-0.6k) & 32.1k &(-0.3k)& 26.4k & (-5.9k) & 23.3K & (-9.0K) \\
    CIFAR-100 & VGG++ & 103.4k & 100.7k &(-2.7k) & 102.8k &(-0.6k)& 93.1k &(-10.2k) & 89.1k & (-14.2) \\
    ImageNet & ResNet50 & 2.33M & 2.27M & (-60.0k) & 2.28M & (-50k) & & & 2.21M & (-120k) \\
\bottomrule
\end{tabular}
\vspace{-0.2cm}
\caption{
Description lengths (in nats) with various techniques added to the baseline model: forward-calibration (FC), label smoothing (LS) and weight standardization (WS).}
\label{tab:ablations}
\end{center}
\end{table}

\begin{figure}[h]
    \centering
    \includegraphics[width=0.48\linewidth]{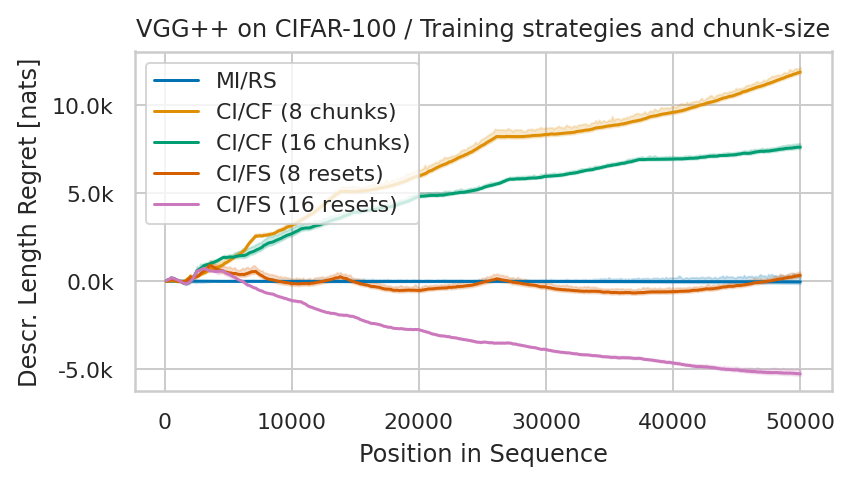} 
    \includegraphics[width=0.47\linewidth]{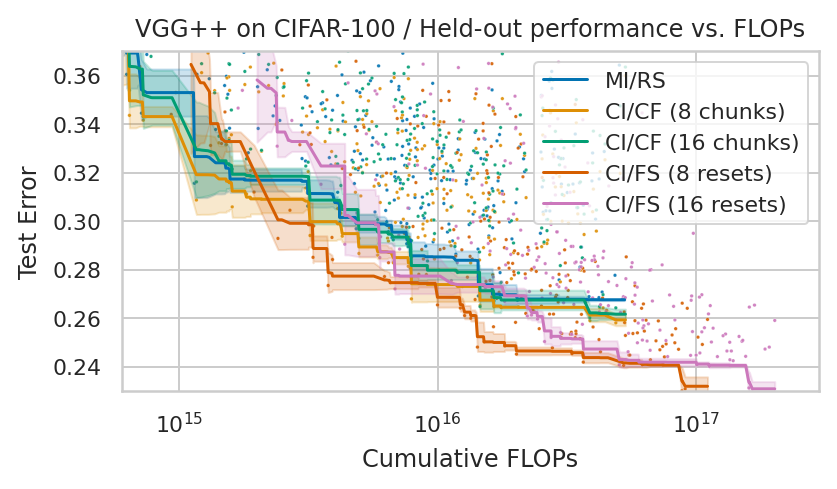} \\
    
    \caption{
    {\bf Left}: Comparing the regret for CI/FS and CI/CF relative to MI/RS for CIFAR-100.
    {\bf Right}: Pareto-front for the test set performance after reaching the end of the sequence 
    }
    \label{fig:cifar100-regret}
\end{figure}

\paragraph{Plasticity and the Warm-starting problem}
We take a closer look at the large-FLOPs regime for CIFAR-100 where CI/FS obtained better results than MI/RS. \autoref{fig:cifar100-regret} (left) shows the regret comparison between the best runs for each training method.
The split positions $s_k$ are clearly visible as kinks in the graph for CI/FS and CI/CF. 
For CI/CF the plot shows that right after fine-tuning, the predictive performance is approximately 
equal to MI/RS.
For CI/FS we see that the regret relative to MI/RS decreases: it has better predictive performance. 
This effect persists even after running extensive additional hyperparameter sweeps. We conjecture that this is a signature of the 
warm-starting problem described in \citep{ash2020warm}. 
\autoref{fig:cifar100-regret} (right) shows the Pareto front for the final test set accuracy of the best  
models (selected by their description length) over FLOPs. We observe a large gap between fine-tuning 
based methods and CI/FS; which is consistent with this hypothesis. Note that the last split-point 
$s_K{=}T$ implies that these models are effectively trained like in a conventional ERM paradigm. 
\autoref{sec:more-experiments} shows test set pareto fronts for a range of datasets. 
Close inspection of various regret plots sometimes reveals the same signatures, 
suggesting that the effect is ubiquitous, but often much less pronounced.
It is plausible that this also contributes to the gap in final test set performance in 
\autoref{tab:all_results}. 

%% file: 05-discussion.tex
\section{Discussion \& Related Work}

Compression based approaches to inference and learning, such as Solomonoff induction \citep{solomonoff1964formal},
Minimum Message Length \citep{Wallace2005} and Minimum Description Length \citep{Rissanen1984,Grunwald2004}, 
have been extensively studied. The Bayesian and variational Bayesian approaches to estimating code
lengths have been popular and, at least outside the field of deep learning, also been very successful.
This line of work includes methods such as AIC and BIC, as well as more sophisticated approximations
to the posterior distribution over model parameters for many kinds of models, including 
neural networks \citep{hinton1993keeping,mackay2003information}.
The description length depends crucially on the quality of these posterior approximations.
Unfortunately, estimating the parameter posterior for modern neural networks is
notoriously hard and an area of active research \citep{izmailov2021bayesian}. 
\cite{Blier2018-rl} demonstrated that much shorter code lengths can be achieved by using 
the prequential approach with standard neural network architectures and optimizers. 
The prequential perspective with the block-wise approximation technique
have been successively used in the context of image-classification \citep{bornschein2020small}, 
natural language processing \citep{DBLP:journals/corr/abs-2003-12298, jin2021causal, perez2021rissanen} 
and causal structure discovery \citep{bornschein2021causal}.

With our work, we aim to simplify and improve prequential code length estimation for neural networks
with insights from the field of continuous learning (CL).
For CL, fine-tuning with rehearsal is considered a strong baseline and 
large number of methods have been proposed to maintain its benefits while minimizing 
the size of the replay buffer \citep{mnih2013playing,delange2021continual}.
We instead propose a method that makes it practical and relatively cheap 
to work with full rehearsal. 
\citet{caccia2022} investigate very similar scenario to ours; largely phrased around how 
long a learner should wait and accumulate data when it is compute constraint; most 
experiments however consider the case without full rehearsal.
He and Lin \citep{helin2020} discuss the close 
relationship between prequential coding and CL, however focus on compressive replay 
instead of accessing previous data.
In general, the literature on CL methods and evaluation metrics is 
too broad to be adequately discussed here \citep{Hadsell2020-fn}. 
However, we would not be surprised if some existing CL methods 
have the  potential to vastly improve prequential MDL computation. 
In this sense, we see our work as a step of bringing CL and 
prequential MDL related works closer together.

Besides of taking inspiration from CL to improve description lengths, 
we believe that the MDL perspective can help guide future CL research:
Much of the CL literature is rooted in an ERM-based perspective
and thus considers distinct training, validation and test sets at different stages during learning. 
This has potentially played a role in the proliferation of many different evaluation metrics and training protocols used in CL research \citep{Hadsell2020-fn}. 
This heterogeneity of the evaluation metrics makes it challenging to compare results.
Forward-transfer is a commonly used evaluation metric and it is closely related to a prequential evaluation. 
However, it is often computed on separate held-out sets after training and does not take 
sample-efficiency into account as prequential MDL does.
Recent work has embraced forward-validation in terms of average prediction-error on future training data as a evaluation metric, 
thus bringing it closer to a prequential MDL based evaluation but without discussing the connection \citep{Cai2021AA, lin2021clear}. 
MDL has been conceived and analyzed as a theoretically well motivated objective when 
dealing with non-stationary data; without reference to the practical difficulties of 
actually performing parameter estimates. 
It is a challenging objective even for learners 
without constraints on compute and (replay-) memory.

\section{Conclusions \& Limitations}

In this paper, we compare different approaches to computing prequential description lengths while considering their  computational cost.
We find that continuous-learning inspired mini-batch incremental training with full rehearsal is the preferred method over a wide range of computational budgets.
We introduce two techniques that either improve results, or make it practical to apply full rehearsal even for very large data.
As a result we report description lengths for a range of standard data sets that improve significantly over previously reported figures.
Some of our results show that the warmstarting-effect %
can have a negative effect on the performance of continual fine-tuning.
We also generally observe a gap in the final held-out performance after training models 
within the prequental framework vs. training models with the ERM-based approach. 
We are not aware of any technique to reliably mitigate the impact. It would be an important topic to tackle for future works.
Furthermore, learning rate schedules such as cosine decay are popular and effective techniques for achieving better performance in ERM-based training. 
It is not obvious how to best translate them to the prequential learning scenario as the learner constantly receives new data.

%% file: 0X-supplement.tex
\section{Pitfalls of using test splits}
\label{sec:splitting}

Test set contamination is an increasingly grave unsolved problem that becomes more pressing as machine learning 
research moves from curated academic benchmark data to large user provided or web-scraped data. 
In the recent GPT-3 paper on large scale language modeling the authors dedicate multiple pages in the 
Appendix to discuss the issue~\citep{brown2020language}. 
The paper details the method used to detect training-/test set duplicates and presents experiments and the conclusion 
that their particular model and evaluation approach is relatively insensitive to contamination. 
Indeed, some of the contamination was only discovered after the model had been trained,
and the model could not be retrained with decontaminated data due to the prohibitive (computational) cost.
This prompted some follow-up discussion on how validation and test sets should be created 
and decontaminated.

In \citep{sogaard2021we} the authors show that depending on the exact splitting procedure significantly 
different results can be obtained. The discussion lays out arguments in favour of either using 
deliberately biased splits, or forward-validating on future data. This work refers 
to prior work \citep{gorman2019we}, which also described inconsistent results between 
different test sets, and between the random split, but at large came to a different conclusion.

To the best of our knowledge these issues have been primarily discussed in the NLP community.
We mention them to point out that validation on held-out data can not blindly be regarded 
as the gold-standard for model evaluation when, at times, the procedure to create those splits
is subject to intense discussion. Instead, careful consideration will be necessary in order to
avoid misleading conclusions based on inappropriate validation and test sets.

Test set contamination is not only an increasingly important problem but also in a certain sense an unsolvable one,
because it is ill-defined. At what threshold for example should a document that verbatimly cites a training-set document be removed from the test-set? 
When are two images scraped from the internet essentially the same?

On the other side: Test set contamination is no problem when using any of the MDL variants for model evaluation. 
For NML and Bayes this is true because they use model complexity and marginal predictions for regularization instead 
of held-out sets to avoid overfitting. But our limited ability to work with good parameter posterior estimates and the 
benign overfitting of overparametrized NNs make it largely non-applicable to deep learning. 
For prequential MDL it is true, because the model is evaluated and hence has to perform well already the first time a (potentially duplicate) data item appears.

For instance, assume the whole photo data base contains three or more (approximate) copies of each photo.
If we randomly split off 10\% as the test set, then the train set contains nearly all test set items.
With heldout-validation a pure memorizer without any generalization capacity will perform nearly 
perfectly on the test set, but will fail miserably in practice on any newly taken photo.
On the other hand, NML and marginal Bayes punish a memorizer due to its huge complexity, 
and prequential MDL due to the inability to predict (well) the first occurrence of each item.
Test set evaluation \emph{is} empirically sound for \emph{i.i.d.} data,
therefore the failure of this paradigm must be attributed to the non-iid nature of data with duplicates \cite{Hutter:22exiid}.

For a more detailed discussion, see the FAQ in \cite{Hutter:06hprize}.

\section{Experimental Details}
\label{sec:experiment-details}
\label{sec:hyperparameters}

\subsection{Forward Calibration}
\label{sec:forward-calibration}

We train the neural network with standard cross-entropy loss and the categorical prediction-head 
$p = \softmax(h)$, where $h$ is a vector of logits generated by the neural network.
When using the model for evaluation on next-step data or on test set examples, we replace it with 
a calibrated prediction head $p = \softmax(\softplus(\beta) \, h)$, where $\beta$ is a scalar temperature 
parameter \citep{guo2017}.
We use {\em forward-calibration} to optimize the calibration parameter $\beta$: each new batch of examples from the 
training sequence is used first for evaluation, then used to perform a single gradient step to optimize $\beta$, 
and finally used for a training to optimizer $\prm$ and placed into the replay-buffer. 
At the beginning of training we initialize $\beta$ to $\beta_0 = \log(\exp(1) - 1) \approx 0.5416$, such that $\softplus(\beta_0) = 1$.
We use the same optimizer as for the model parameters $\prm$, however scale the learning rate by a factor of $\sqrt{K}$,
where $K$ is the number of training steps per evaluation step forward (i.e., number replay-streams plus one). 
Calibration is computationally almost free: the forward-pass for the calibration gradient step can be shared with the 
evaluation forward-pass, and backward-propagation for $\beta$ ends shortly after the softmax layer.

\subsection{MNIST, EMNIST and RGB-MNIST Hyperparameters}

We run experiments with various model architectures: 
a) MLP with 1, 2 or 3 hidden layers, 512 units each, with dropout, and ReLU non-linearities,
b) LeNet \citep{lecun1989} and 
c) the VGG inspired architecture used by \citep{Blier2018-rl}, which we call MnistNet here.

We use the same hyperparameter and sampling intervals for all three model architectures. 
For the pareto front experiment we extend the number of replay-streams
range to $10 \dots 100$; otherwise we use the same intervals.

\begin{center}
\begin{tabular}{lcl}
    \toprule
    Parameter & Distribution & Values / Interval \\
    \midrule
    Number of replay-streams & log-uniform & 25 \dots 100 \\
    Learning rate & log-uniform &  1e-4 \dots 3e-3 \\
    AdamW $\epsilon$ & log-uniform & 1e-4 \dots 1 \\
    EMA step size (Polyak averaging) & log-uniform & 1e-3 \dots 1e-1 \\ 
    Weight decay & log-uniform & 1e-4 \dots 1. \\
    Batch size & fixed & 32 \\
    Label smoothing & fixed & 0.001 \\
\end{tabular}
\end{center}

\subsection{CIFAR-10 and CIFAR-100 with VGG++, and WideResNet Hyperparameters}

We use the same hyperparmater sampling intervals for all model architectures.
Following \cite{ZagoruykoKomodakis2016} we use only minimal augmentations during training: 
Images will be horizontally flipped with a probability of 0.5. For the pareto-front plots 
we extend the {\em number of epochs} (= number of replay-streams) range to $10 \dots 100$ and

\begin{center}
\begin{tabular}{lcl}
    \toprule
    Parameter & Distribution & Values / Interval \\
    \midrule
    Number of replay-streams & log-uniform & 25 \dots 100 \\
    Learning rate & log-uniform &  1e-4 \dots 3e-3 \\
    AdamW $\epsilon$ & log-uniform & 1e-4 \dots 1 \\
    EMA step size (Polyak averaging) & log-uniform & 1e-3 \dots 1e-1 \\ 
    Weight decay & log-uniform & 1e-4 \dots 1e-1 \\
    Batch size & uniform & \{32, 64, 128\} \\
    Label smoothing & fixed & 0.01 \\
\end{tabular}
\end{center}

\subsection{ImageNet with VGG++ and ResNets Hyperparameters}

We use randaugment for data augmentation and the same hyperparameter intervals for all experiments. For Pareto front experiments we additionally 
scale the architecture size (number of channels throughout) from $\sfrac{1}{4} \times$ to $4 \times$ and extend the number of replay-streams interval to 
$10 \dots 100$.

\begin{center}
\begin{tabular}{lcl}
    \toprule
    Parameter & Distribution & Values / Interval \\
    \midrule
    Number of replay-streams & log-uniform & 25 \dots 100 \\
    Learning rate & log-uniform &  1e-4 \dots 3e-3 \\
    AdamW $\epsilon$ & log-uniform & 1e-4 \dots 1 \\
    EMA step size (Polyak averaging) & log-uniform & 1e-3 \dots 1e-2 \\ 
    Weight decay & log-uniform & 1e-4 \dots 1. \\
    Batch size & fixed & 512 \\
    Label smoothing & fixed & 0.01 \\
\end{tabular}
\end{center}

\subsection{CLOC with ResNets Hyperparameters}
\label{sec:cloc-hypers}

We use randaugment for data augmentation and the same hyperparameter intervals for all experiments.

\begin{center}
\begin{tabular}{lcl}
    \toprule
    Parameter & Distribution & Values / Interval \\
    \midrule
    Number of replay-streams & log-uniform & 10 \dots 25 \\
    Learning rate & log-uniform &  1e-4 \dots 3e-3 \\
    EMA step size (Polyak averaging) & log-uniform & 1e-3 \dots 1e-2 \\ 
    Batch size & fixed & 128 \\
    Label smoothing & fixed & 0.1 \\
\end{tabular}
\end{center}

\section{Additional Empirical Results}
\label{sec:more-experiments}

\subsection{RGB-MNIST Feature Selection}
\label{sec:rgbmnist-appendix}

To illustrate that the model-selection property demonstrated in \autoref{sec:rgbmnist} is a robust 
property of (prequential) MDL, we here present a suite of results on 3 different model architectures:
1) am MLP with 2 hidden layers, 512 units each; 
2) the LeNet architecture from \citep{lecun1989}; 
and 3) the much higher capacity and better tuned convnet from \citep{Blier2018-rl}. 

In all cases we run the same hyperprameter sweep detailed in \autoref{sec:hyperparameters}. We show the regret 
plot for the best performing model in each sweep relative to the model conditioned only on the R channel. 
We see pMDL reliably and with confidence determines that the model conditioned on R only is the most 
appropriate one for this data; and we see that the small-data performance at the beginning of the sequence 
is crucial as the relative generalization performance becomes more similar as the length of the sequence increases:

\begin{center}
{\bf MLP Regret Plot} \\
\includegraphics[width=0.99\linewidth]{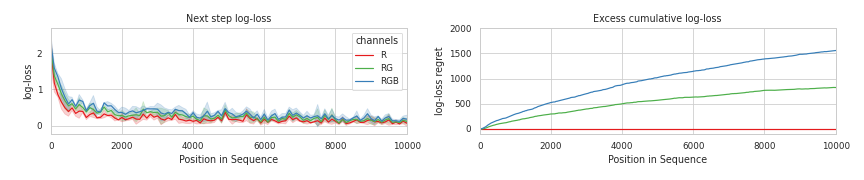} \\
{\bf LeNet Regret Plot} \\
\includegraphics[width=0.99\linewidth]{plots/rgbmnist-insta-regret.png} \\  %
{\bf MnistNet Regret Plot} \\
\includegraphics[width=0.99\linewidth]{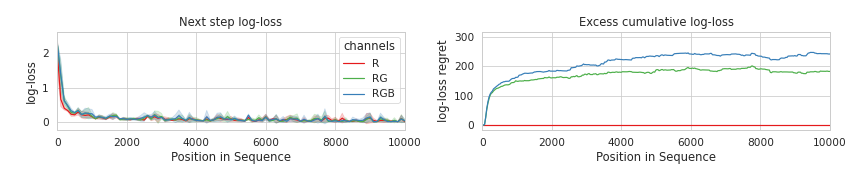} \\
\end{center}

\newpage
\subsection{Pareto Fronts}
\label{fig:pareto-detail}

{\bf Left:} Description lengths as function of compute resources consumed for a selection 
of datasets and models. 
{\bf Right:} Test set performance after reaching the end of the training data. 
We generally observe a gap between models that train from scratch (CI/FS) and continual finetuning 
methods. The gap is very pronounced for CIFAR-100. 
We do not observe a systematic gap between MI/RS and CI/CF.

\begin{center}
\bf MNIST with LeNet \\
\includegraphics[width=\linewidth]{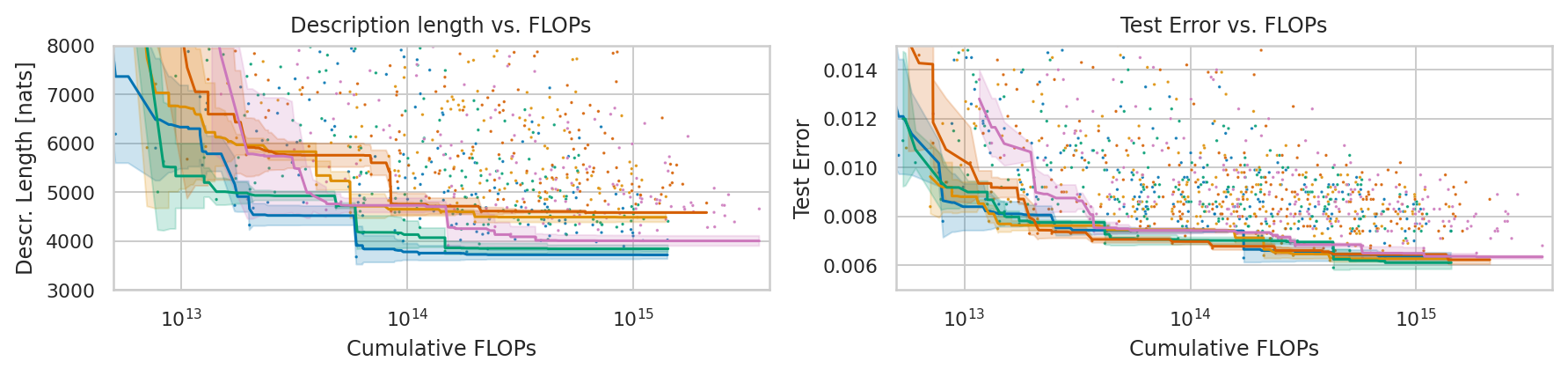}\\
\bf EMNIST with LeNet \\
\includegraphics[width=\linewidth]{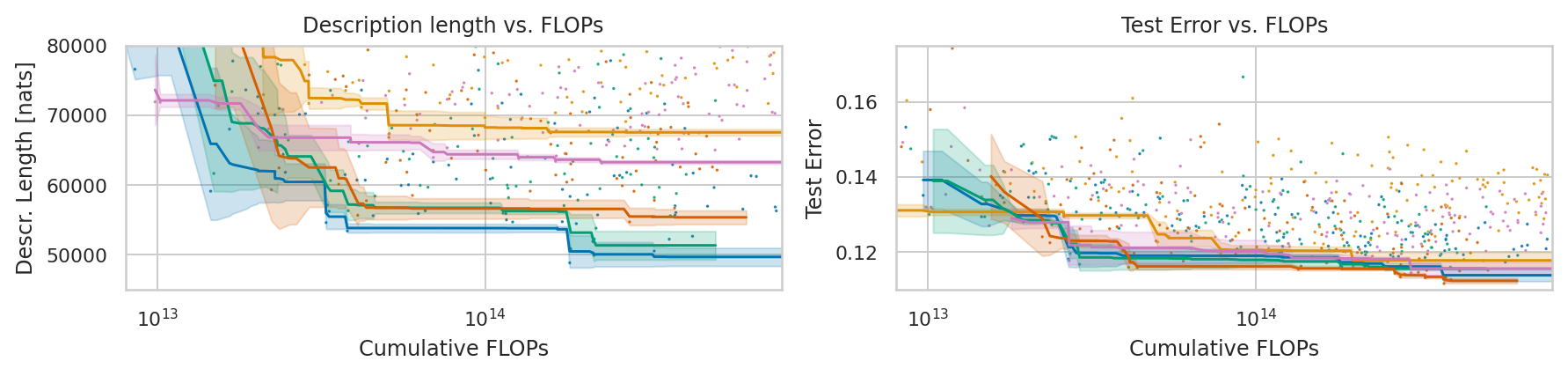}\\
\bf CIFAR-10 with VGG++ \\
\includegraphics[width=\linewidth]{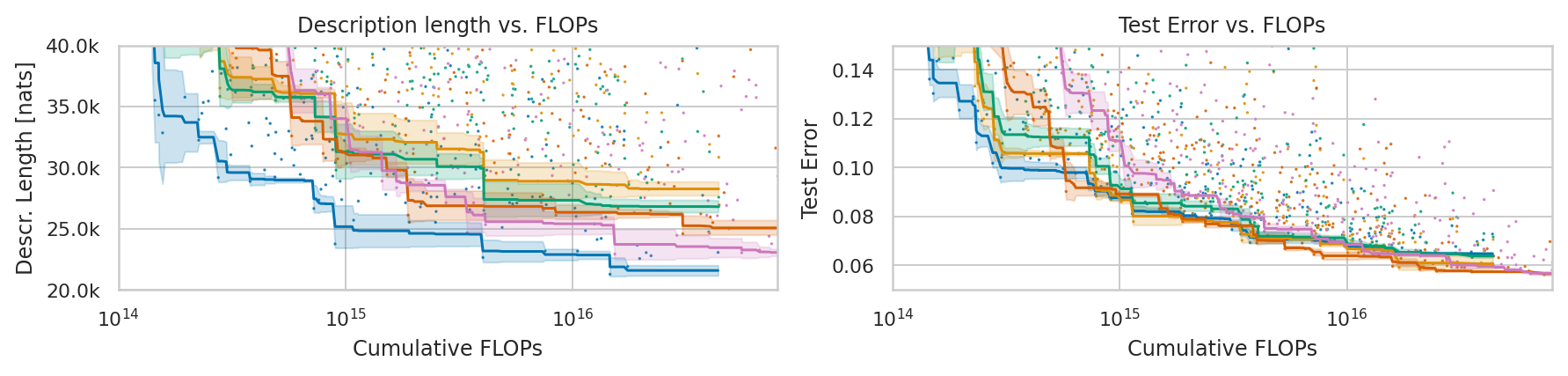}\\
\bf CIFAR-100 with VGG++ \\
\includegraphics[width=\linewidth]{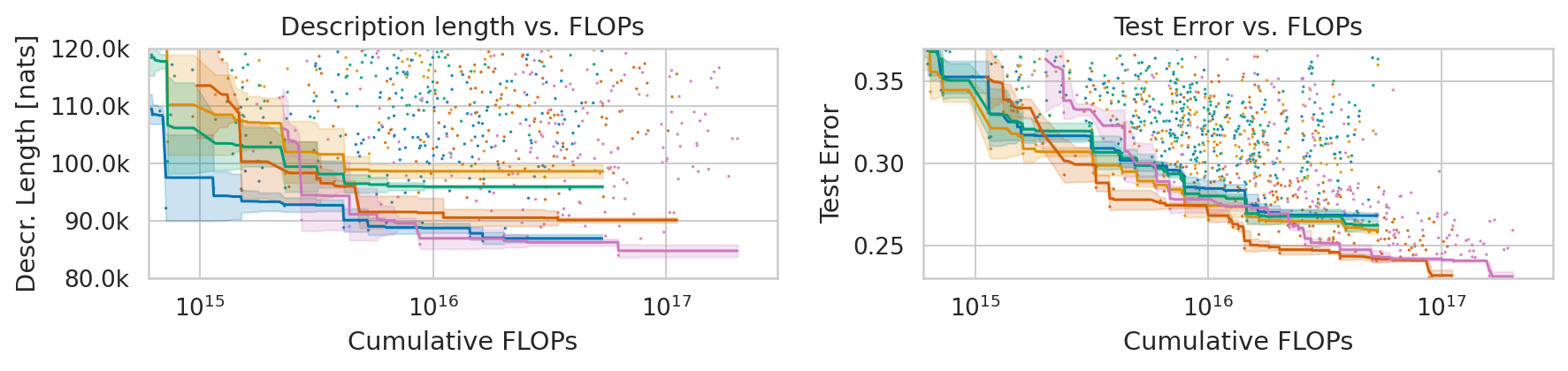}\\
\bf ImageNet with ResNet-50 \\
\includegraphics[width=\linewidth]{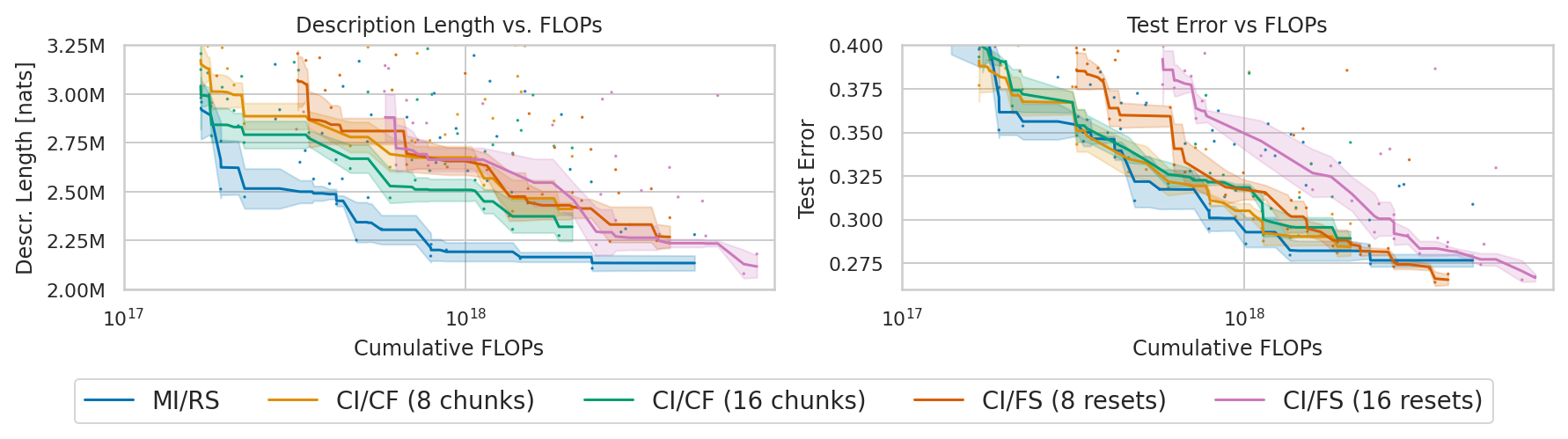}\\
\end{center}

\subsection{Model Depth Ablation}
\label{sec:model-depth-ablations}

\includegraphics[width=0.48 \linewidth]{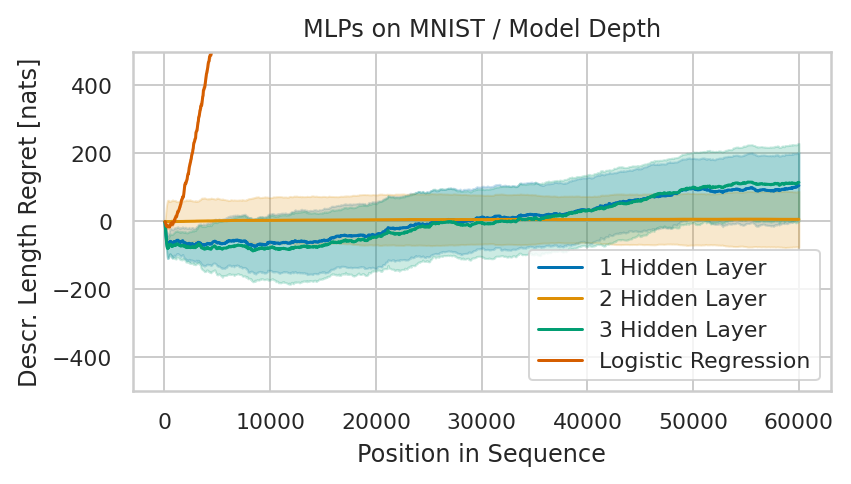}
\includegraphics[width=0.48 \linewidth]{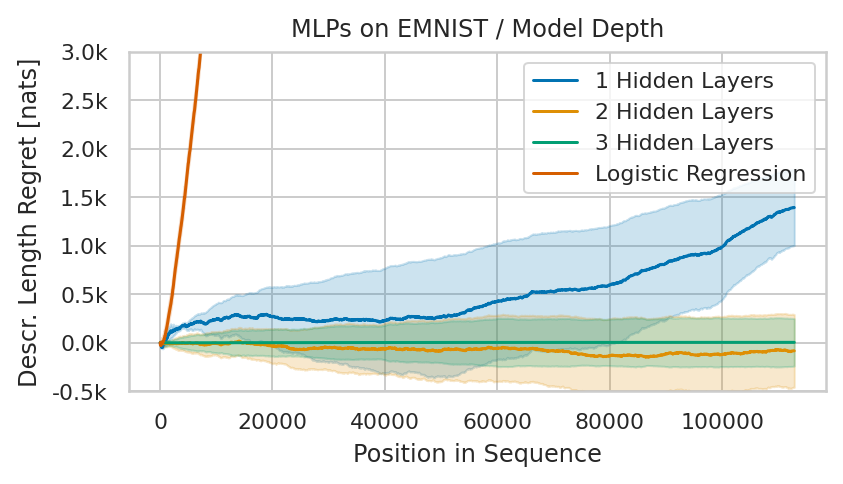}\\

\subsection{Weight-Standardization Ablation}
\label{sec:ws-ablations}

\includegraphics[width=0.48 \linewidth]{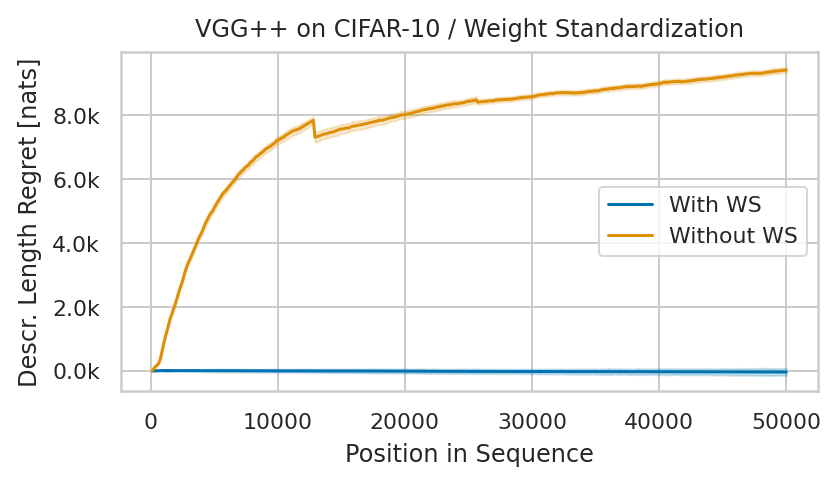}
\includegraphics[width=0.48 \linewidth]{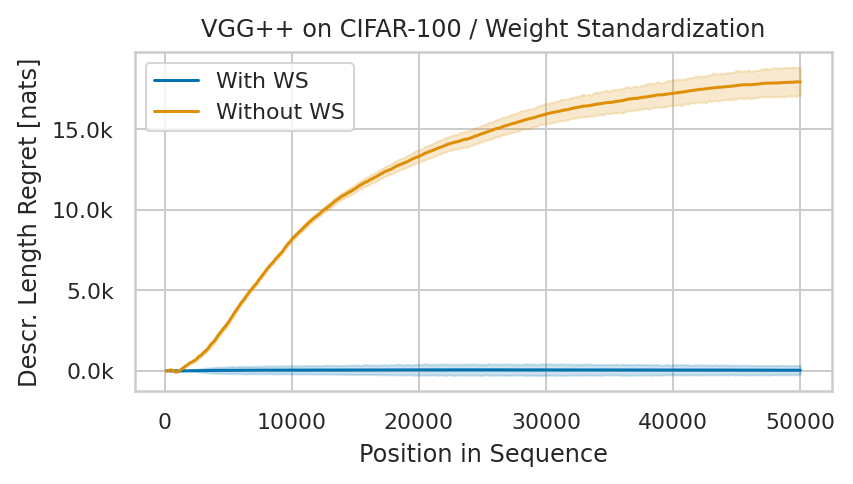} \\

\subsection{Number of Replay Streams}
\label{sec:replay-streams-ablations}

We run ablations for different number of replay-streams ($K$ in \Cref{alg:replay-stream}) on CIFAR-100 and ImageNet. 
The results on ImageNet show that performing too much replay (choosing K too large) can be harmful.

\includegraphics[width=0.48 \linewidth]{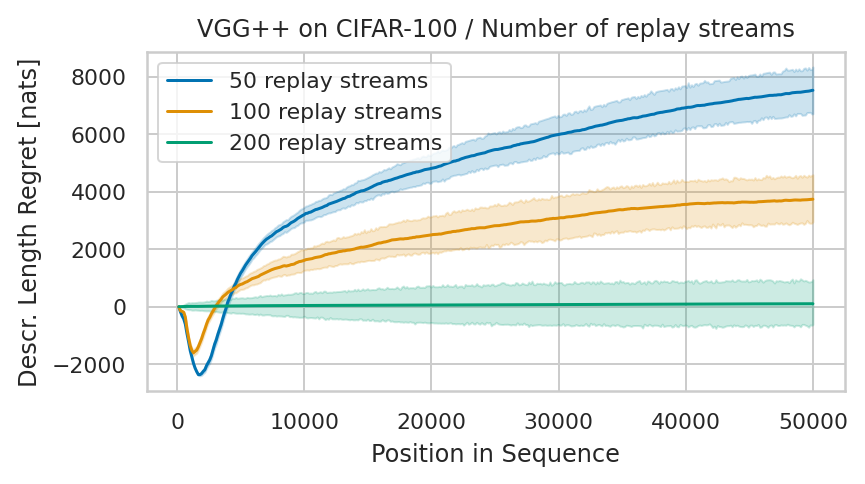}
\includegraphics[width=0.48 \linewidth]{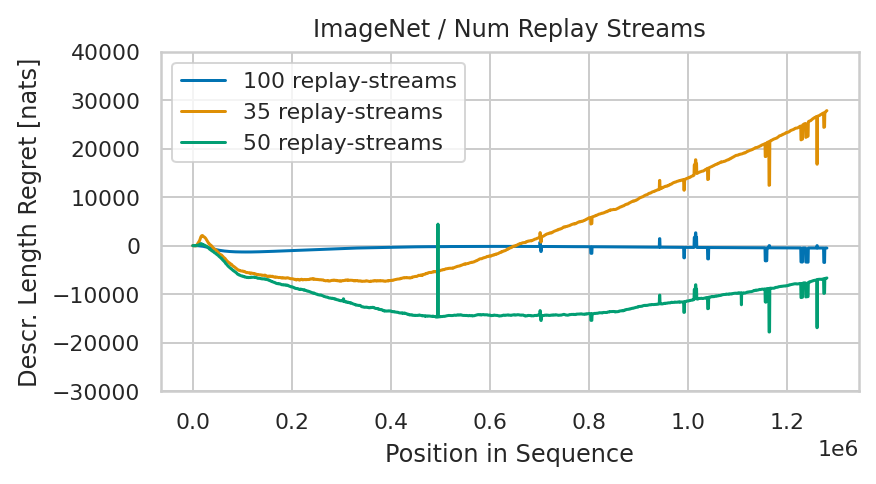}

\subsection{Scale and shape of the Replay Distribution}
\label{sec:replay-distributions}

Besides of uniform replay for replay-stream training we experimented
with alternative replay distributions.  
Exponentially decaying replay ($p_\text{replay}(a) = \lambda \exp(-\lambda a)$) is popular 
in continual-learning with reservoir sampled replay buffers, but lead to significant longer
description lengths and suboptimal results in our experiments.

As an compromise between uniform replay and exponential decay we experimented with 
long-tailed Pareto distributions: 
$p_\text{replay}(a) = \frac{\alpha}{(\sfrac{a}{\lambda})ˆ{(\alpha + 1)}}$.
The scale parameter $\lambda$ determines to what extend old examples are replayed
and we use a fixed shape parameter $\alpha = \log_4 5 \approx 1.16$ for all out experiments: 

\includegraphics[width=0.48 \linewidth]{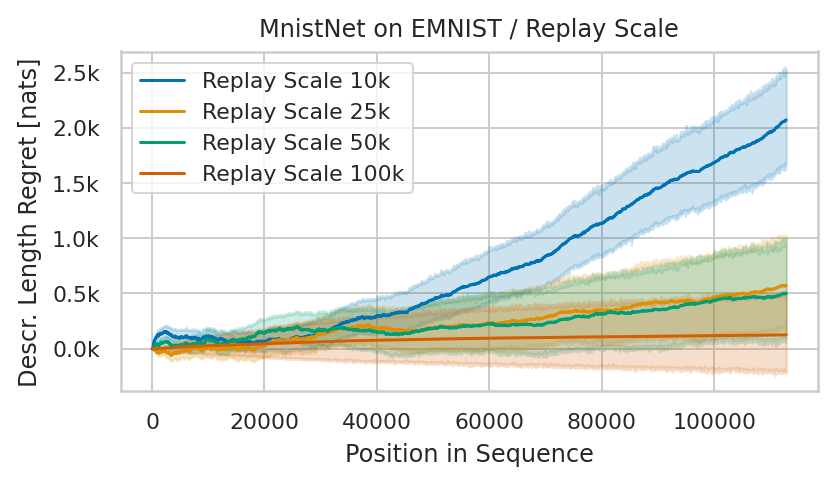}
\includegraphics[width=0.48 \linewidth]{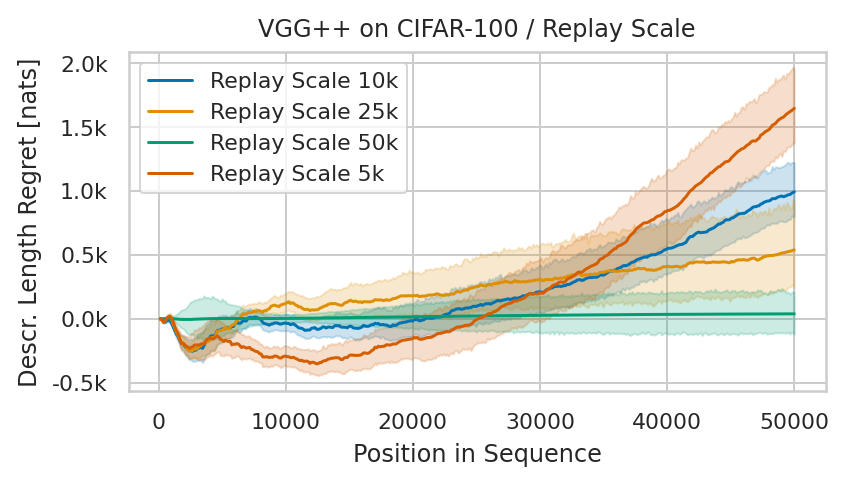} \\

Our experiments suggest that flat, almost uniform replay distributions lead to the shortest
description lengths for the sequences from i.i.d. sources.

\subsection{Non-stationary sequence: Continual geo-localization (CLOC)}
\label{sec:cloc}

In this study we concentrate on data sequences from stationary (i.i.d) data sources to ensure we 
can compare our results with established ERM based training. 
However, one major benefit of MDL over ERM is that it is well defined for individual and non-stationary sequences. 
We here present empirical results on the CLOC data introduced in \citep{Cai2021AA}. 
CLOC is a large scale, chronologically sorted image classification sequence consisting of $\approx 39M$ images 
with their geolocation as a categorical label. For $\approx 5\%$ of the 
images we either received download errors, or the downloaded content could 
not be decoded as image data. In total we were able to obtain 37,093,769 labeled images.
Figure 2 in \citep{Cai2021AA} shows that the data is strongly non-stationary; and that 
traditional ERM training, which treats the data as i.i.d., results in a held-out error rate 
between 80 and 90\%.

In \citep{Cai2021AA} the authors reserve the first 2M ($5\%$) datapoints for 
pre-training the model statically before entering the online-phase, split-off
$1\%$ of the data throughout for validation purposes and only evaluate the 
images from the first new album in each mini-batch instead of all images.
Their online-learning task and metric are thus in spirit similar to the prequential 
MDL problem we are tackling here, in detail however sufficiently different that the 
quantitative results are not directly comparable.
Following \citep{Cai2021AA} we use a standard ResNet50 architecture and run experiments 
with replay buffers and replay streams. The hyperparameter sampling space is detailed in 
Appendix \ref{sec:cloc-hypers}. Note that we use a batch size of 128 because small 
batch-sizes allow the model to more quickly adapt to changes in the data distribution; 
and a relatively low number of 10-25 replay-steps / replay-streams to limit the total 
computational cost due to the size of the data.

We observe that heavily recency biased replay leads to significantly improved results. 
For replay streams we experimented with uniform replay, replay with exponentially decaying 
priority and with replay governed by a heavy tailed Pareto distribution (see \ref{sec:replay-distributions}).
From these the heavy tailed Pareto replay leads to the shortest description lengths. Even more than 
in the i.i.d. case, forward-calibration plays an crucial role to obtain good results.

\includegraphics[width=0.48 \linewidth]{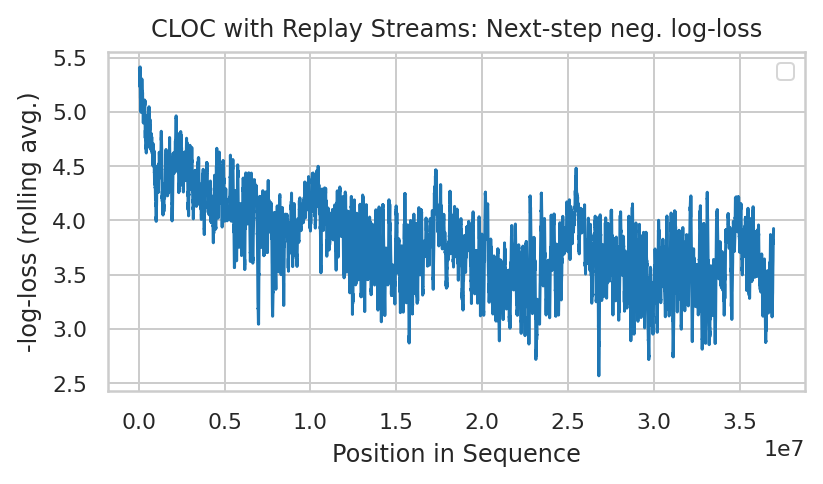}
\includegraphics[width=0.48 \linewidth]{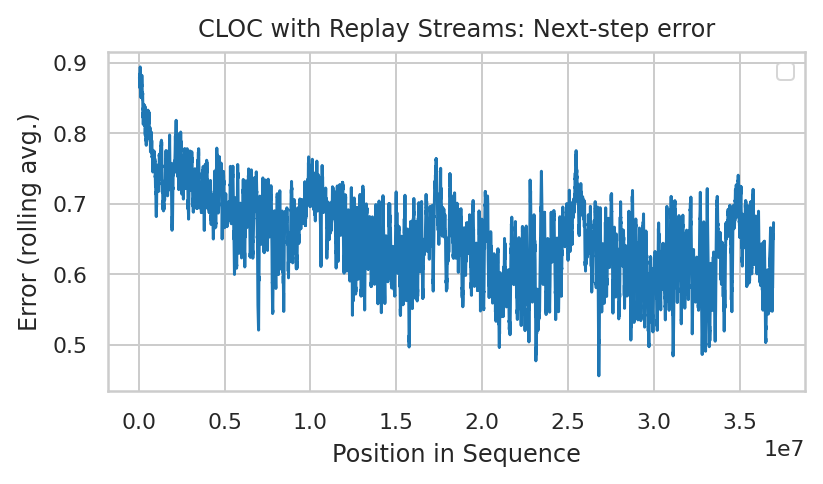} \\

\paragraph{Next-step performance}
We plot the 5k rolling average next-step performance of the best performing model from our 
random hyperparameter sweep ($K{=}22$ Pareto distributed replay streams with a scale of 180).
The model obtains a description length of 1.33G nats and a next-step error-rate of typically $60 \pm 10\%$.

\paragraph{Ablations}
{\bf Left:} We show the description length regret for the best replay-buffer training run relative
to the best replay-stream training run. Both experiments consumed about $5 \times 10^{18}$ FLOPs.
{\bf Right:} Regret for replay-streams without forward calibration ralative to replay streams with forward calibration:

\includegraphics[width=0.48 \linewidth]{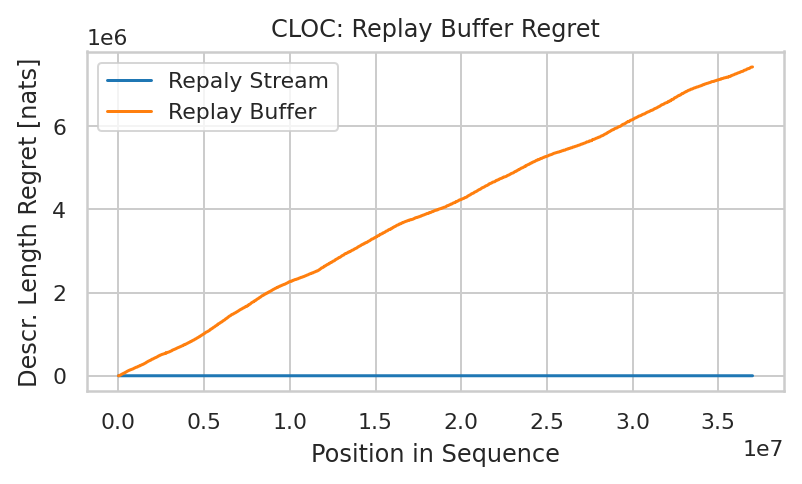}
\includegraphics[width=0.48 \linewidth]{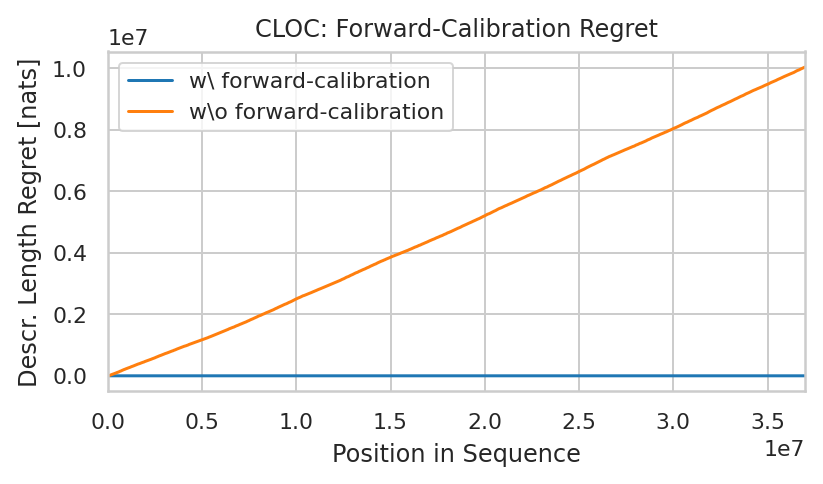} \\